# Object Detection and Motion Planning for Automated Welding of Tubular Joints

Syeda Mariam Ahmed[1], Yan Zhi Tan[2], Gim Hee Lee[1,2], Chee Meng Chew[1], and Chee Khiang Pang[2]

*Abstract*— Automatic welding of tubular TKY joints is an important and challenging task for the marine and offshore industry. In this paper, a framework for tubular joint detection and motion planning is proposed. The pose of the real tubular joint is detected using RGB-D sensors, which is used to obtain a real-to-virtual mapping for positioning the workpiece in a virtual environment. For motion planning, a Bi-directional Transition-based Rapidly exploring Random Tree (BiTRRT) algorithm is used to generate trajectories for reaching the desired goals. The complete framework is verified with experiments, and the results show that the robot welding torch is able to transit without collision to desired goals which are close to the tubular joint.

## I. INTRODUCTION

Deployment of intelligent robots for industrial automation is a growing trend, particularly for the marine and offshore industry. Due to the availability of robotic manipulators with specialized sensors, simple welding tasks can be performed with high accuracy and repeatability. However, welding of tubular TKY joints is still carried out manually. This is due to the complex geometry which requires multi-pass welding. An example of a tubular joint is shown in Fig. 1. Conventionally, the welding joint is divided into several sections. The welding deformation. As such, it is necessary to sense the environment for motion planning.

The major aspects of welding automation include object detection/seam finding and motion planning. Seam finding refers to adjusting the welding torch of the manipulator to the correct pose in relation to the welding groove. Optical sensor based smart systems for seam finding are offered by an increasing number of companies. Examples include Servo Robot Inc (Robo-Find, Power-Trac, i-CUBE) and Meta Vision Systems (Laser Pilot). For these systems, seam finding is carried out using highly accurate laser sensors and CCD cameras. There are also tactile solutions which make use of the filler material or the gas nozzle as a mechanical sensor [1]. However, tactile solutions can only be used to determine the pose of simple welding geometries. In addition, most of these solutions are designed for close range sensing (within a few centimeters).

For motion planning, the common industrial solutions [2, 3] are online and walk-through programming. Using online programming, the waypoints of the trajectory are recorded as the user guides the robot using a teaching pendant.

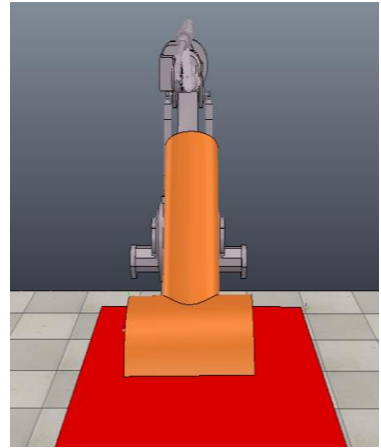

Fig 1. Robotic manipulator and tubular joint in V-REP simulation

On the other hand, Kinetiq teaching [5] is an example of walk-though programming, where the robot welding torch is guided by hand to the desired location. This method is effective for simple joints which require single pass welding and do not involve complicated, large welding seams [4].

An alternative approach will be to develop an offline simulation environment which is a scaled replica of the real robot environment [6, 7]. Examples of commercial softwares which allow development of such workcells include RobotStudio and Robotmaster. In these softwares, the robot kinematic model is imported from a library along with 3D CAD models of the workpieces. One of the significant research efforts to implement offline programming was by Pan et al., where an Automated OffLine Programming (AOLP) software was developed for welding path planning using high degree of freedom robots [8]. In AOLP, the trajectories for transitions between welding tasks are computed using Probabilistic RoadMap (PRM).

The Rapidly-exploring Random Trees (RRT) algorithm [9] is also commonly used for automatic trajectory generation in high-order workspace for robotic manipulators. The initial point is regarded as the root of a tree, and the workspace is explored using random sampling. Feasible waypoints are added as a node of the tree which expands towards the desired goal. The RRT connect planner is one of the early adaptations, where bi-directional trees expand alternatively from initial and goal nodes [10]. In order to generate a cost-optimal trajectory, the RRT* [11] algorithm is proposed.

Syeda Mariam Ahmed[1] and Chee Meng Chew[1] are with the Department of Mechanical Engineering, National University of Singapore (email: mpesyed@nus.edu.sg).

Yan Zhi Tan[2] and Chee Khiang Pang[2] are with the Department of Electrical Engineering, National University of Singapore.

Gim Hee Lee[1,2] is with the Department of Mechanical Engineering and the Department of Electrical Engineering, National University of Singapore.

Other variations include the anytime RRT [12] and lazy RRT [13] algorithms. The latest extensions include informed-RRT* [14] and wrapping-based informed RRT* [15], where the sampling space is limited to a hyperellipsoid.

In this paper, a framework is proposed for tubular welding joint detection and motion planning. The proposed framework uses open source libraries without the need for licensed software. Besides, the simulation environment is easily updated based on real world information by an object detection component of the framework. In addition, a comparative study of motion planners for transitioning between welding tasks is carried out.

## II. FRAMEWORK FOR OBJECT DETECTION AND MOTION PLANNING

Welding automation for tubular joints can be divided into two phases. For the first phase, object detection and motion planning for transitions between welding tasks is carried out. Welding trajectory planning and laser-based seam tracking is the main objective of the second phase. In this paper, the focus is on the implementation of the first phase. The proposed framework is shown in Fig. 2, which consists of two main components:

1) Object detection: Perception of the *real* workshop environment is carried out using two RGB-D sensors as shown in Fig. 2. A *virtual* environment is created using V-REP [16] by importing CAD models of the robot, tubular joints and a workbench. This environment can also serve as a user interface, such as allowing the user to change the pose of the welding torch. The pointclouds from the sensors are used to perform a *real-to-virtual* world mapping. This mapping determines the misalignment in pose between the real and virtual workpieces, and can be used to update the simulation for all subsequent motion planning tasks.
2) Motion planning: Bi-directional Transition-based Rapidly exploring Random Tree (BiTRRT) algorithm [17] is used for generating the trajectories for transitions between the welding segments. Optimality for the generated trajectories is based on minimizing the Integral of the Cost (IC) along a path. The planned trajectories are visualized in the virtual environment before uploading to the robot controller.

In the following subsections, a more detailed description of the proposed framework will be provided.

### A. Object Detection

Perception of environment is an essential prerequisite for welding automation. In this section, the RGB-D setup, pointcloud generation from CAD models, preprocessing of pointclouds, and registration for determining real-to-virtual mapping are described with reference to Fig. 2.

The two RGB-D sensors are mounted on tripod stands and placed perpendicularly to each other at known distances from the workbench as shown in Fig. 2. The sensor positions are fixed and the environment can be reconstructed using simple transformations. Additional RGB-D sensor can always be deployed in order to obtain a complete pointcloud of the workpiece with all key features included.

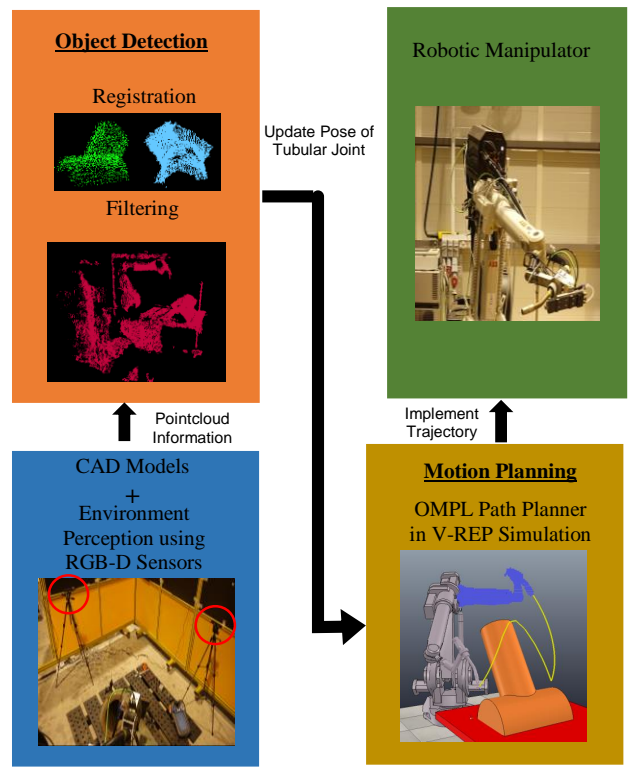

Fig 2. Overall framework comprising of object detection and motion planning for automating the welding of a tubular joint.

A point cloud from the V-REP environment is generated using CAD models. A CAD model can be used to generate a STereoLithography (STL) file which can subsequently be converted to a Polygon File Format (PLY). The PLY model is primarily a sparse pointcloud comprising of only the vertices of the CAD model. A denser pointcloud is generated by *raytracing* using the Point Cloud Library (PCL) [18]. In PCL, raytracing is implemented by creating a spherical truncated icosahedron around the PLY model. A virtual camera pointed towards the origin is placed at each vertex of the icosahedron. Depth information of the model is used to simulate the input of a depth sensor, and multiple snapshots from each pose are combined to create the dense pointcloud.

In order to perform real-to-virtual world mapping, registration between the pointclouds is required. As the focus is on the tubular joint, segmentation is carried out prior to registration, and the workbench in the real and virtual worlds are pre-aligned. The alignment is carried out by assigning a work frame to the edge of the workbench. The real position of this work frame is recorded by jogging the robot to the physical edge and reading the coordinates from the teaching pendant. This position is used to update the workbench placement in V-REP. After segmentation and workbench alignment, noise filtering using the Difference of Normal (DoN) operator [19] is applied to the pointcloud obtained from the RGB-D sensors in order to improve registration accuracy. The DoN operator can be defined as

$$\Delta_{\hat{n}}(p, r_1, r_2) = \frac{\hat{n}(p, r_1) - \hat{n}(p, r_2)}{2}, r_1, r_2 \in \Re, r_1 < r_2 \quad (1)$$

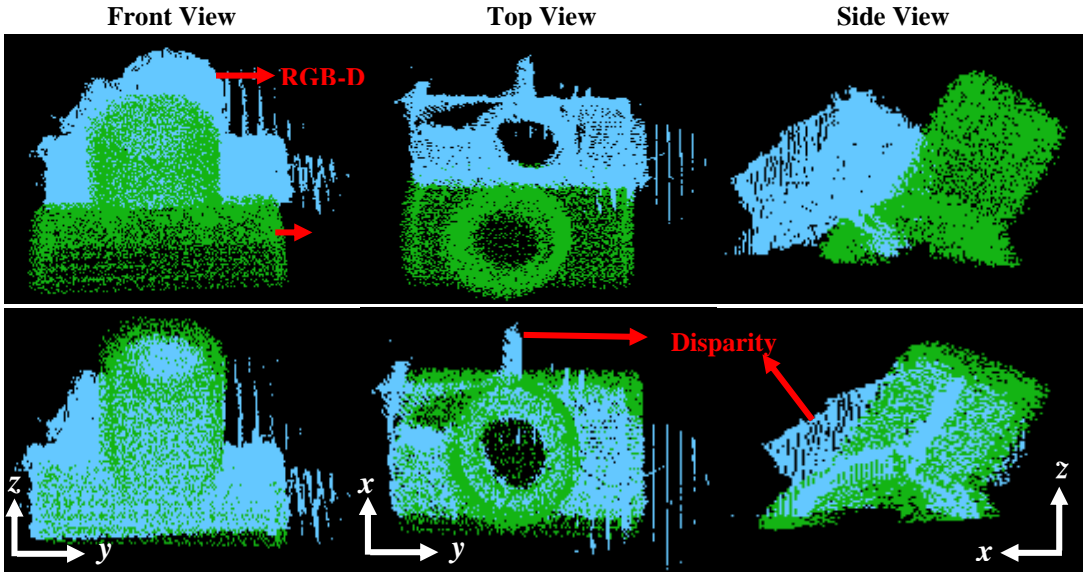

Fig 3. Pointclouds from real and virtual environment before (top) and after (bottom) ICP registration

where $\hat{n}(p, r_1)$ and $\hat{n}(p, r_2)$ represents the estimated surface normals at a point p considering support radii r1 and r2, respectively. The result of applying the DoN operator to the pointcloud is a vector map of $\Delta_{\hat{n}}$, where $\|\Delta_{\hat{n}}\| \in [0,1]$. The vector map represents the difference in surface normal considering a varying support radius, and noise can be filtered out by specifying a threshold for $\|\Delta_{\hat{n}}\|$. Registration is performed using the point-to-point Iterative Closest Point (ICP) algorithm in PCL. ICP employs brute force approach which consists of two main steps to align two point clouds. First, the nearest neighbor in the target point cloud is determined for every point from the source pointcloud. Next, the optimal transformation which minimizes the sum of squared distances between the corresponding points is determined. The nearest neighbor search in PCL is implemented using Fast Library for Approximate Nearest Neighbors (FLANN) [23]. FLANN consists of a collection of algorithms optimized for fast nearest neighbor search in large datasets. From the ICP algorithm, a transformation matrix is obtained which is used to update the pose of the tubular joint in the virtual environment.

### B. Motion Planner for Transition between Welding Tasks

Motion planning is carried out after the pose of the tubular joint is updated. The welding task for the tubular joint is divided into several sections, and the welding sections are switched after every pass in order to avoid joint deformation. In this paper, the objective of motion planning is to ensure a safe transition for the robotic manipulator between these sections. As such, drastic configuration changes should be avoided in order to prevent damage to the equipment. In addition, another motivation for automated offline motion planning is to avoid singularities.

The V-REP environment is recently integrated with the Open Motion Planning Library (OMPL) [20] through APIs. As such, the motion planning task can be easily defined for the planner. A kinematic model for the robotic manipulator is set up in the environment, and the user is required to define the elements of the kinematic chain. The V-REP environment provides the flexibility to choose between the pseudo-inverse and the Damped Least Squares (DLS) method for inverse kinematics computation. With the virtual environment and the robot kinematic model, various path planners can be analyzed in order to determine the most appropriate planner for the task.

For our welding tasks, the BiTRRT [17] algorithm is used to generate the transition trajectories. The BiTRRT algorithm is a combination of RRT and the stochastic-optimization method which is used to compute the global minima in complex spaces. The expansion of BiTRRT is similar to the RRT connect planner [10]. At each iteration, the new node is subjected to a transition test based on the Metropolis criterion. Using the Metropolis criterion [22], the cost of the new configuration is compared with the cost of its parent configuration. If the criterion is satisfied by the new node, the node is connected to the nearest neighbor in the other tree. The planned trajectories are visualized in the virtual environment and uploaded to the robot controller.

### III. EXPERIMENTS AND RESULTS

In this section, the proposed framework is implemented on a 6 degree-of-freedom robotic manipulator from ABB in our workshop. The workpiece considered is shown in Fig. 1, which is a typical tubular joint. Besides verifying the object detection component by aligning the real and virtual tubular joints, the robustness of the ICP algorithm is explored by considering additional position and orientation offsets. For optimal motion planning, cost functions for measuring the divergence of the robot pose from the goal throughout the trajectory are introduced.

### A. Tubular Joint Alignment Using Registration

The pointclouds of the real environment is obtained using two RGB-D sensors as shown in Fig. 2. In order to prevent interference between the RGB-D sensors, recording is carried out using one sensor at a time. The pointcloud is segmented using the thresholding technique in Cartesian space in order to retain only the tubular joint. For a change in workpiece position within the setup, an identical threshold can be used for segmentation as the sensors remain at fixed positions. The DoN operator for noise filtering is implemented using $r_1 = 5$, $r_2 = 50$ with the threshold for $\|\Delta_{\hat{n}}\|$ set to 0.1. The threshold value is chosen such that features of the TKY joint are preserved. The segmented and filtered pointcloud of the tubular joint is as shown in blue in Fig. 3.

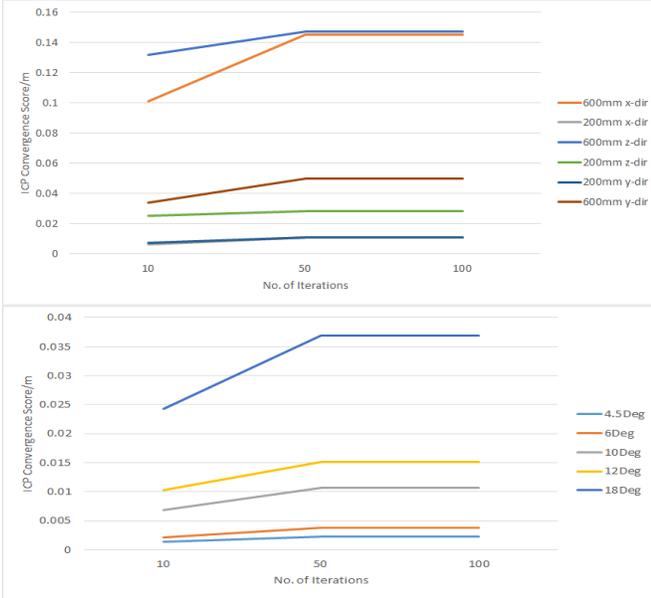

Fig 4. ICP convergence score considering various position (top) and orientation (bottom) offsets between the real and virtual environments.

Registration is carried out using ICP from PCL, and the results after ten iterations are shown in Fig .3. The pointclouds before and after registration are shown in the top and bottom rows of Fig. 3, respectively. The green pointclouds in Fig. 3 represent the pointclouds from the CAD model. The transformation matrix for real-to-virtual mapping is mainly a translation of 200 mm in the *x*-direction, and the virtual environment is updated accordingly.

From Fig. 3, disparity between the pointclouds from the real and virtual environment can be observed. The green pointclouds from the CAD model do not have fixtures as compared to the pointclouds of the actual tubular joint. The fixtures are not included in the CAD model for ICP registration as these fixtures are not precisely positioned for every joint. Besides, the position of the vertical pipe may vary within an allowable range.

A series of simulations were also conducted to evaluate the robustness of the ICP algorithm in the presence of such disparity. The position offsets considered are up to 600 mm in each of the *x*, *y* and *z* directions. Similarly, the orientation offsets are up to 18° about the *z*-axis. The algorithm converges to a score which is slightly higher than the minimum ICP convergence score as shown in Fig. 4. This may be due to the points belonging to the fixtures which are not aligned with any points from CAD model. The results show that the ICP convergence score increases significantly only when an offset of 600 mm in the *x* and *z*-direction or a rotation of 18° about the *z*-axis is given. This range is higher than the working range of commercial optical and tactile sensors-based systems for seam-finding. Besides, such level of misalignment in the *z*-axis is unlikely as the workbenches are pre-aligned.

### B. Cost Analysis of Bidirectional Transition-Based RRT

Collision-free and cost-optimal paths are required for transitioning between the welding tasks. The IC criterion is used to assess the quality of the trajectories generated by the motion planner. A discrete approximation of the integral cost $c_p$ for a path $\pi$ can be defined as

$$c_p(\pi) = \frac{L}{n}\sum_{k=1}^{n} c(\pi(k)), \quad (2)$$

where $L$ represents the length of the path, and $n$ denotes the number of subdivisions along the path. Besides, $c$ in (2) represents a continuous differentiable cost function that can be replaced with either $c_{pos}$ and $c_{orient}$ which are defined as

$$c_{pos} = \|X_k - X_{target}\|, \quad (3)$$

$$c_{orient} = \min\{\|(q_k - q_{target})\|, \|(q_k + q_{target})\|\}. \quad (4)$$

In (3) and (4), $X_k$, $q_k$, $X_{target}$, and $q_{target}$ represent the current position, current orientation, goal position, and goal orientation vectors considering the tip of the welding torch as the tool center position, respectively. The Euler angle representation is used for the orientation vectors. Using the IC criterion, the divergence of the path from the goal is evaluated at every intermediate node as compared to the maximal or average cost criteria.

Paths for reaching two goal poses are generated by the BiTRRT algorithm over fifteen trials, and the IC values computed using the two cost functions $c_{pos}$ and $c_{orient}$ are shown in Fig. 5. The IC values considering $c_{pos}$ are observed to be low for multiple trials, in particular for the trajectory to reach Goal 1 as shown in Fig. 6.

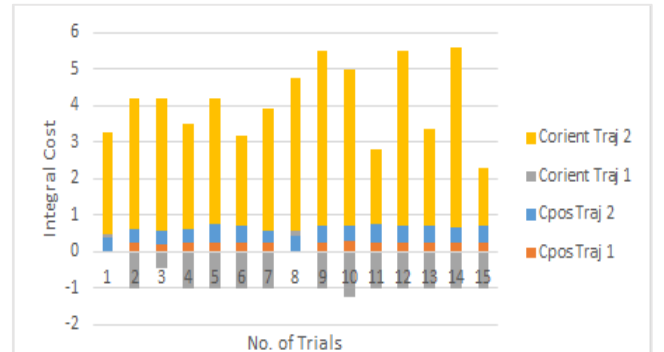

Fig 5. Evaluation of $c_p$ for BiTRRT planner using $c_{pos}$ and $c_{orient}$ functions.

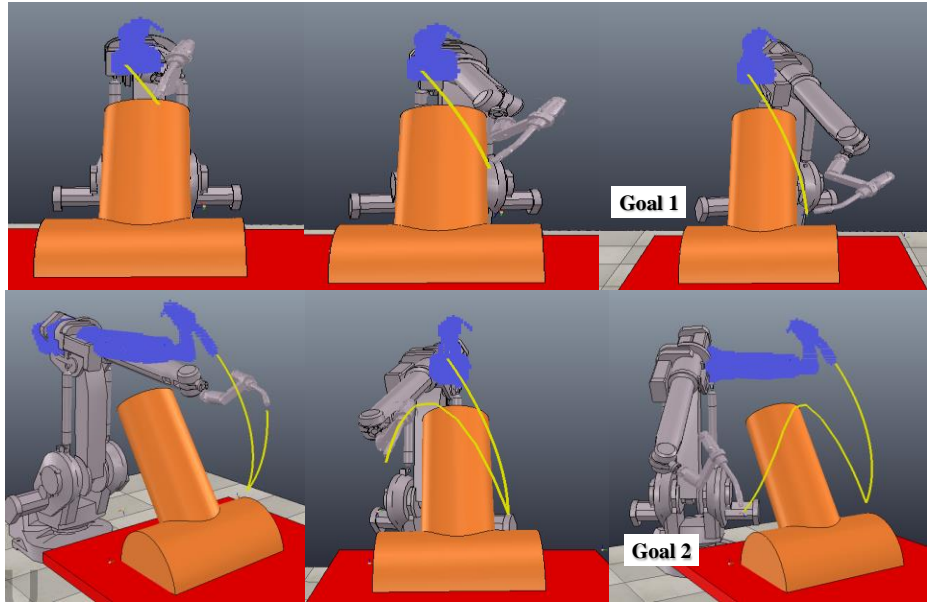

Fig 6. Simulation verification of proposed framework. (top) Trajectory generated to reach Goal 1. (bottom) Trajectory to transit from Goal 1 to Goal 2.

The variation in IC values is larger for $c_{orient}$ due to differences in the avoidance of the fixture by the robotic manipulator. However, the optimal paths generated by the planner are consistent as indicated by the IC values which remain within a close range for both trajectories.

Trajectories generated by the BiTRRT algorithm for the two goal poses and implementation results on the robotic manipulator are shown in Fig. 6. The first trajectory from the robot home position to Goal 1 is shown in the top row, where Goal 1 represents the approach point for the first welding task. The intermediate trajectory to transit from Goal 1 to Goal 2 for the next welding task is shown in the bottom row of Fig. 6. Similar results are obtained when the trajectories are implemented on the real welding robot. Various motion planners will be compared in the next section based on plots of cost functions $c_{pos}$ and $c_{orient}$ in order to verify the effectiveness of the BiTRRT algorithm.

### C. Comparative Study of Motion Planners

Optimal motion planning for a robotic manipulator can be described by the divergence of the intermediate nodes within the trajectory. Besides, the generated path should avoid drastic configuration changes in order to prevent damage to the welding and monitoring equipment as mentioned in Section II-B. A comparative analysis of planners available in OMPL is performed to analyze the cost functions $c_{pos}$ and $c_{orient}$ as defined in (3) and (4). Each planner is evaluated considering the two goals specified in Section III-B, and the following algorithms are compared:
  a) *Bidirectional Transition Based RRT (BiTRRT)*,
  b) *Probabilistic Roadmap Star (PRMStar)*,
  c) *Lower-Bound Tree RRT (LBT-RRT) [21]*,
  d) *RRT Connect, and*
  e) *RRT Star.*

The plots of $c_{pos}$ and $c_{orient}$ over the time taken to reach Goal 1 are shown in Figs. 7 (a) and (c). The plots of $c_{pos}$ and $c_{orient}$ for Goal 2 are similarly shown in Figs. 7(b) and (d). The area under the curves is largest for Goal 1 using the LBT-RRT algorithm. For Goal 2, LBT-RRT and PRMStar results in the largest area for $c_{orient}$ and $c_{pos}$, respectively. The area is larger due to the divergence of the intermediate states from the goal, which results in a longer trajectory for the robotic manipulator. On the other hand, the BiTRRT algorithm is able to generate a path with a minimum area under the curve. As the rest of the algorithms have comparable performance, the BiTRRT is chosen due to the consistency observed in IC values for paths generated over multiple iterations as shown in Fig. 5.

### IV. CONCLUSION AND FUTURE WORK

A framework for tubular TKY joint detection and motion planning is proposed in this paper. The proposed framework involves environment perception to align workpieces in the virtual world based on real world positioning, which is a common problem for deploying intelligent robotic manipulators in the industry. This issue is addressed by capturing a pointcloud of the real world using RGB-D sensors and generating a similar pointcloud from CAD models. A real-to-virtual world mapping is obtained through a series of pre-processing steps followed by registration. Subsequently, a BiTRRT motion planner is used to plan trajectories for transitions between the welding tasks. Comparative evaluation of different motion planners supported by the OMPL shows that the BiTRRT algorithm can consistently generate low cost paths. Besides, the implemented trajectories for a tubular joint are shown.

Future works include considering the RGB-D sensor to be mounted on the robotic manipulator. This will improve the scalability of the proposed framework and facilitate deployment in industries.

ACKNOWLEDGMENT

The authors thank the National Research Foundation, Keppel

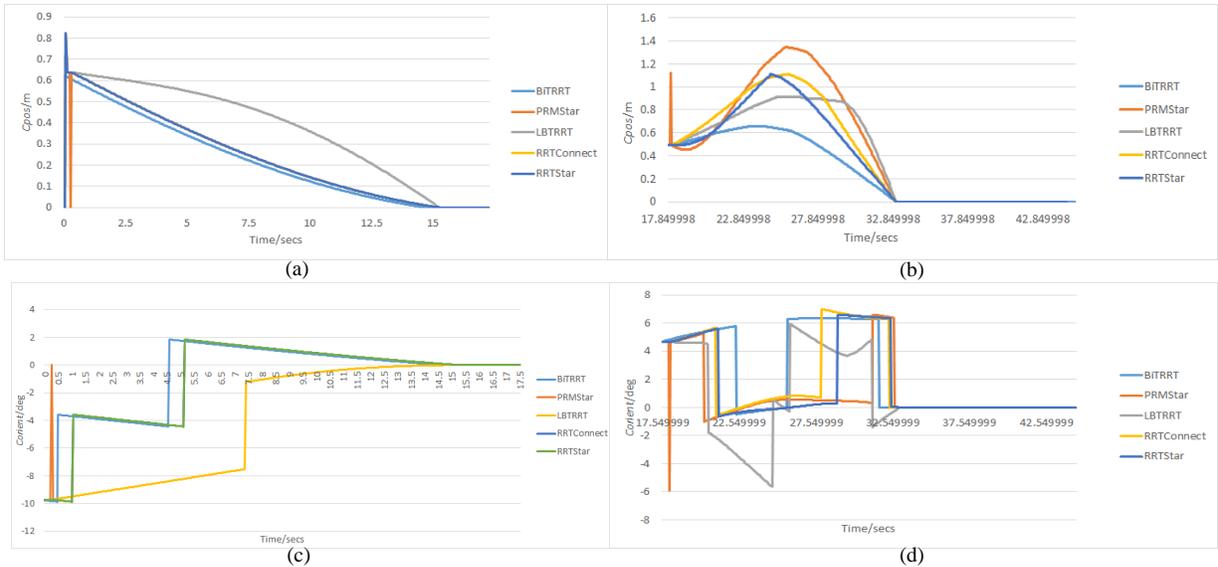

Fig 7. Comparison of motion planners using plots of cpos and corient functions. (a) cpos values for Goal 1. (b) cpos values for Goal 2. (c) corient values for Goal 1. (d) corient values for Goal